# Asymptotic constant-factor approximation algorithm for the Traveling Salesperson Problem for Dubins' vehicle

Ketan Savla  Emilio Frazzoli  Francesco Bullo

*Abstract*— This article proposes the first known algorithm that achieves a constant-factor approximation of the minimum length tour for a Dubins' vehicle through $n$ points on the plane. By Dubins' vehicle, we mean a vehicle constrained to move at constant speed along paths with bounded curvature without reversing direction. For this version of the classic Traveling Salesperson Problem, our algorithm closes the gap between previously established lower and upper bounds; the achievable performance is of order $n^{2/3}$.

## I. INTRODUCTION

The Traveling Salesperson Problem (TSP) with its variations is one of the most widely known combinatorial optimization problems. While extensively studied in the literature, these problems continue to attract great interest from a wide range of fields, including Operations Research, Mathematics and Computer Science. The Euclidean TSP (ETSP) [1], [2] is formulated as follows: given a finite point set $P$ in $\mathbb{R}^2$, find the minimum-length closed path through all points in $P$. It is quite natural to formulate this problem in context of Dubins' vehicle, i.e., a non-holonomic vehicle that is constrained to move along paths of bounded curvature, without reversing direction. The focus of this article is the analysis of the TSP for Dubins' vehicle; we shall refer to it as DTSP.

Exact algorithms, heuristics as well as polynomial-time constant factor approximation algorithms are available for the Euclidean TSP, see [3], [4], [5]. It is known that non-metric versions of the TSP are, in general, not approximable in polynomial time [6]. Furthermore, unlike most other variations of the TSP, it is believed that the DTSP cannot be formulated as a problem on a finite-dimensional graph, thus preventing the use of well-established tools in combinatorial optimization. On the other hand, it is reasonable to expect that exploiting the geometric structure of Dubins' paths one can gain insight into the nature of the solution, and possibly provide polynomial-time approximation algorithms.

A fairly complete picture is available for the minimum-time point-to-point path planning problem for Dubins' vehicle, see [7] and [8]. However, the DTSP seems not to have been studied as extensively. In [9], some results for the worst case tours of DTSP were provided. A lower bound on the expected cost of a stochastic DTSP visiting randomly generated points was provided in [10]. Here, we shall

Ketan Savla and Francesco Bullo are with the Center for Control, Dynamical Systems and Computation, University of California at Santa Barbara, ketansavla@umail.ucsb.edu, bullo@engineering.ucsb.edu

Emilio Frazzoli is with the Mechanical and Aerospace Engineering Department, University of California at Los Angeles, frazzoli@ucla.edu

specifically concentrate on the case when the target points in the environment are generated stochastically according to a uniform distribution. We shall refer to such a problem as stochastic DTSP. In this context the first algorithm with asymptotic sub-linear cost was proposed in [11]; an improved algorithm was proposed in [12].

The motivation to study the DTSP arises in robotics and uninhabited aerial vehicles (UAVs) applications, e.g., see [13], [14], [15]. In particular, we envision applying our algorithm to the setting of an UAV monitoring a collection of spatially distributed points of interest. Additionally, from a purely scientific viewpoint, it appears to be of general interest to bring together the work on Dubins' vehicle and that on TSP. Some concrete results along these lines have been obtained in [9] and in [11] where an algorithm to guarantee sub-linear cost for the stochastic DTSP was proposed. UAV applications also motivate us to study the Dynamic Traveling Repairperson Problem (DTRP), in which the aerial vehicle is required to visit a dynamically changing set of targets. This problem was introduced by Bertsimas and van Ryzin in [16] and then decentralized policies achieving the same performances were proposed in [13]. However, as with the TSP, the study of DTRP in context of Dubins' vehicle has eluded attention from the research community.

The contributions of this article are twofold. First, we propose an algorithm for the stochastic DTSP through a point set $P$, called the RECURSIVE BEAD-TILING ALGORITHM, based on a geometric tiling of the plane, tuned to the Dubins' vehicle dynamics, and a strategy for the vehicle to service targets from each tile. Second, we obtain an upper bound on the stochastic performance of the proposed algorithm and thus also establish a similar bound on the stochastic DTSP. The upper bound on the performance of the RECURSIVE BEAD-TILING ALGORITHM belongs to $O(n^{2/3})$ with high probability, and it is known that the lower bound on the achievable performance belongs to $\Omega(n^{2/3})$. The algorithm we introduce in this article is the first known algorithm providing a provable constant-factor approximation to the DTSP optimal solution.

*Notation*

Here we collect some concepts that will be required in the later sections. A *Dubins' vehicle* is a planar vehicle that is constrained to move along paths of bounded curvature, without reversing direction and maintaining a constant speed. Accordingly, we define a *feasible curve for Dubins' vehicle* or a *Dubins' path*, as a curve that is twice differentiable almost everywhere, and such that the magnitude of its curvature is bounded above by $1/\rho$, where $\rho > 0$ is the minimum turn radius.

Let $P = \{p_1, \ldots, p_n\}$ be a set of $n$ points in a compact region $\mathcal{Q} \subset \mathbb{R}^2$ and $\mathcal{P}_n$ be the collection of all point sets $P \subset \mathcal{Q}$ with cardinality $n$. Let $\text{ETSP}(P)$ denote the cost of the Euclidean TSP over $P$, i.e., the length of the shortest closed path through all points in $P$. Correspondingly, let $\text{DTSP}_\rho(P)$ denote the cost of the Dubins' TSP over $P$, i.e., the length of the shortest closed Dubins' path through all points in $P$. In what follows, $\rho \in \mathbb{R}_+$ is take constant, and we study the dependence of $\text{DTSP}_\rho : \mathcal{P}_n \to \mathbb{R}_+$ on $n$.

For $f, g : \mathbb{N} \to \mathbb{R}$, we say that $f \in O(g)$ (respectively, $f \in \Omega(g)$) if there exist $N_0 \in \mathbb{N}$ and $k \in \mathbb{R}_+$ such that $|f(N)| \le k|g(N)|$ for all $N \ge N_0$ (respectively, $|f(N)| \ge k|g(N)|$ for all $N \ge N_0$). If $f \in O(g)$ and $f \in \Omega(g)$, then we use the notation $f \in \Theta(g)$.

## II. THE STOCHASTIC DTSP

In [9], a simple heuristics, the ALTERNATING ALGORITHM for the Dubins' TSP for a given point set was proposed. The length of tour generated by this algorithm was also characterized and it was shown that it belongs to $\Omega(\sqrt{n})$ and $O(n)$. It was also shown that this simple policy performs well when the points to be visited by the tour are chosen in an adversarial manner. However, it is reasonable to argue that this algorithm might not perform very well when dealing with a random distribution of the target points. In particular, one can expect that when $n$ points are chosen randomly, the cost of the DTSP increases sub-linearly with $n$, i.e., that the average length of the path between two points decreases as $n$ increases. In this section, we consider the scenario when $n$ target points are stochastically generated in $\mathcal{Q}$ according to a uniform distribution. A novel algorithm, the BEAD-TILING ALGORITHM was proposed in [11] to service these points in such a way that its tour length grew sub-linearly with the number of points asymptotically with high probability, where an event is said to occur with high probability if the probability of its occurence approaches 1 as $n \to +\infty$. Here, we present a novel version of this strategy in the form of the RECURSIVE BEAD-TILING ALGORITHM and characterize its performance.

We assume that the environment $\mathcal{Q}$ is a rectangle of width $W$ and height $H$; different choices for the shape of $\mathcal{Q}$ affect our conclusions only by a constant. In what follows we select a reference frame whose two axes are parallel to the sides of $\mathcal{Q}$. Let $n$ target points be generated stochastically according to uniform distribution in the region $\mathcal{Q}$. Let $\mathcal{P} = (p_1, \ldots, p_n)$ be the locations of these target points.

### A. A lower bound

First, we summarize a result from [10], that provides a lower bound on the expected length of the stochastic DTSP.

*Theorem 2.1: (Lower bound on stochastic DTSP)* For all $\rho > 0$, the expected cost of a stochastic DTSP visiting a set $P$ of $n$ uniformly-randomly-generated points in $\mathcal{Q}$, $\text{E}[\text{DTSP}_\rho(P)]$ belongs to $\Omega(n^{2/3})$.

### B. A constructive upper bound

In [11], a novel algorithm, the BEAD-TILING ALGORITHM, to compute Dubins' path through a point set in the region $\mathcal{Q}$ was proposed. In this section, we design the RECURSIVE BEAD-TILING ALGORITHM based on the ideas of the BEAD-TILING ALGORITHM. We will show that the proposed algorithm provides a constant factor approximation to the optimal DTSP with high probability. We start by describing some useful geometric objects.

*1) The basic geometric construction:* Consider two points $p_- = (-l, 0)$ and $p_+ = (l, 0)$ on the plane, with $l \le \rho$, and construct the region $\mathcal{B}_\rho(l)$ as detailed in Figure 1. In

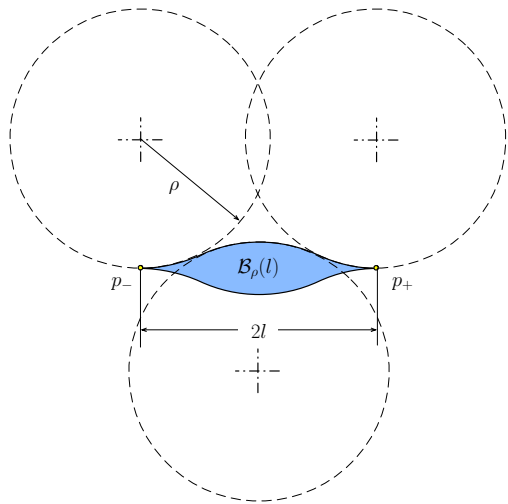

Fig. 1. Construction of the "bead" $\mathcal{B}_\rho(l)$. The figure shows how the upper half of the boundary is constructed, the bottom half is symmetric.

the following, we will refer to such regions as *beads*. The region $\mathcal{B}_\rho(l)$ enjoys the following asymptotic properties as $(l/\rho) \to 0^+$:

(P1) The maximum "thickness" of the region is equal to
$$w(l) = 4\rho\left(1 - \sqrt{1 - \frac{l^2}{4\rho^2}}\right) = \frac{l^2}{2\rho} + o\left(\frac{l^3}{\rho^3}\right).$$

(P2) The area of $\mathcal{B}_\rho(l)$ is equal to
$$\text{Area}[\mathcal{B}_\rho(l)] = lw(l) = \frac{l^3}{2\rho} + o\left(\frac{l^4}{\rho^4}\right).$$

(P3) For any $p \in \mathcal{B}_\rho$, there is at least one Dubins' path $\gamma_p$ through the points $\{p_-, p, p_+\}$, entirely contained within $\mathcal{B}_\rho$, and such that its length is at most
$$\text{Length}(\gamma_p) \le 4\rho \arcsin\left(\frac{l}{2\rho}\right) = 2l + o\left(\frac{l^2}{\rho^2}\right).$$

These facts are verified using elementary planar geometry.

*2) Periodic tiling of the plane:* An additional property of the geometric shape introduced above is that the plane can be periodically tiled by identical copies of $\mathcal{B}_\rho(l)$, for any $l \in (0, 2\rho]$. (Recall that a tiling of the plane is a collection

of sets whose intersection has measure zero and whose union covers the plane.) Let

$$\mu(l) = \frac{\text{Area}[\mathcal{B}_\rho(l)]}{\text{Area}[\mathcal{Q}]}.$$

Consider a bead $B$ entirely contained in $\mathcal{Q}$; the probability that the $i$-th point is sampled in $B$ is equal to $\mu$. Furthermore, the probability that exactly $k$ out of the $n$ points are sampled in $B$ has a binomial distribution, i.e., indicating with $n_B$ the total number of points sampled in $B$,

$$\text{Prob}[n_B = k|n \text{ samples}] = \binom{n}{k} \mu^k (1-\mu)^{n-k}. \quad (1)$$

Choose $\mu$ as a function of $n$, in such a way that $\nu = n\mu(n)$ is a constant. In such a case, the limit for large $n$ of the binomial distribution (1) is the Poisson distribution of mean $\nu$, that is,

$$\lim_{n \to \infty} \text{Prob}[n_B = k|n \text{ samples}] = \frac{\nu^k}{k!} e^{-\nu} \quad (2)$$

*C. The algorithm*

Consider a tiling of the plane such that $\text{Area}[\mathcal{B}_\rho(l)] = WH/(2n)$; in such a case, $\mu = 1/(2n)$, and $\nu = 1/2$. (Note that this implies that $n$ must be large enough that $l < 2\rho$.) Furthermore, the tiling is chosen is such a way that it is aligned with the sides of $\mathcal{Q}$, see Figure 2.

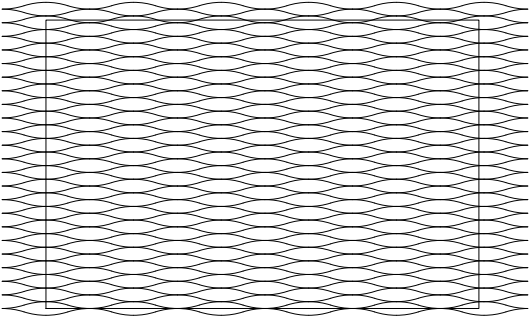

Fig. 2. Sketch of tiling of the region before the first phase of the RECURSIVE BEAD-TILING ALGORITHM.

The proposed algorithm will consist of a sequence of phases; during each of these phases, a Dubins tour (i.e., a closed path with bounded curvature) will be constructed that "sweeps" the set $\mathcal{Q}$.

In the first phase, a Dubins tour is constructed with the following properties:
  (i) it visits all non-empty beads once,
  (ii) it visits all rows[1] in sequence top-to-down, alternating between left-to-right and right-to-left passes, and visiting all non-empty beads in a row,
  (iii) when visiting a non-empty bead, it services at least one target in it.

In order to visit the outstanding targets, a new phase is initiated. In this phase, instead of considering single beads,

[1]A row is a maximal string of beads with non-empty intersection with $\mathcal{Q}$.

we will consider "meta-beads" composed of two beads each, as shown in Figure (3), and proceed in a similar way as the first phase, i.e., a Dubins tour is constructed with the following properties:
  (i) the tour visits all non-empty meta-beads once,
  (ii) it visits all (meta-bead) rows in sequence top-to-down, alternating between left-to-right and right-to-left passes, and visiting all non-empty meta-beads in a row,
  (iii) when visiting a non-empty meta-bead, it services at least one target in it.

This process is iterated at most $\log_2 n + 1$ times, and at each phase meta-beads composed of two neighboring meta-beads from the previous phase are considered; in other words, the meta-beads at the $i$-th phase are composed of $2^{i-1}$ neighboring beads. After the last phase, the leftover targets will be visited using, for example, a greedy strategy, or the Alternating Algorithm.

We have the following result, which we prove using a technique similar to that developed in [17].

*Lemma 2.2:* Let $P \in \mathcal{P}_n$ be uniformly randomly generated in $\mathcal{Q}$. Then, the number of unvisited targets after the last phase of the RECURSIVE BEAD-TILING ALGORITHM over $P$ belongs to $O(\log n)$ with high probability.

*Proof:* Associate a unique identifier to each bead, e.g., integers between 1 and $2n$; call such a set of identifiers $\mathcal{I}$. Let $b(t) \in \mathcal{I}$ be the identifier of the bead in which the $t$-th target is sampled, and let $h(t) \in \mathbb{N}$ be the phase at which the $t$-th target is visited. Without loss of generality, we will assume that if $b(t_1) = b(t_2)$, and $t_1 < t_2$, then $h(t_1) < h(t_2)$.

Indicate with $v_i(t)$ the number of beads that contain unvisited targets at the inception of the $i$-th phase, computed after the insertion of the $t$-th target. Furthermore, let $m_i$ be the number of $i$-th phase meta-beads (i.e., meta-beads containing $2^{i-1}$ neighboring beads) with a non-empty intersection with $\mathcal{Q}$. Clearly, $v_i(t) \leq v_i(n)$, $m_i \leq 2m_{i+1}$, and $v_1(n) \leq n \leq m_1/2$ with certainty.

The $t$-th target will not be visited during the first phase if it is sampled in a bead that already contains other targets. In other words,

$$\Pr[h(t) \geq 2|v_1(t)] = \frac{v_1(t)}{m_1} \leq \frac{v_1(n)}{2n} \leq \frac{1}{2}.$$

Similarly, the $t$-th target will not be visited during the $i$-th phase if (i) it has not been visited before the $i$-th pass, and (ii) it belongs to a meta-bead that already contains other targets not visited before the $i$-th phase:

$$\Pr[h(t) \geq i+1|(v_i(t-1), v_{i-1}(t-1), v_1(t-1))]$$
$$= \Pr[h(t) \geq i+1|h(t) \geq i, v_i(t-1)] \cdot$$
$$\Pr[h(t) \geq i|(v_{i-1}(t-1), \ldots, v_1(t-1))]$$
$$\leq \frac{v_i(t-1)}{m_i} \Pr[h(t) \geq i|(v_{i-1}(t-1), \ldots, v_1(t-1))]$$
$$= \prod_{j=1}^{i} \frac{v_j(t-1)}{m_j} \leq \prod_{j=1}^{i} \frac{2^{j-1} v_j(n)}{2n} = \left(\frac{2^{\frac{i-3}{2}}}{n}\right)^i \prod_{j=1}^{i} v_j(n).$$
$$(3)$$

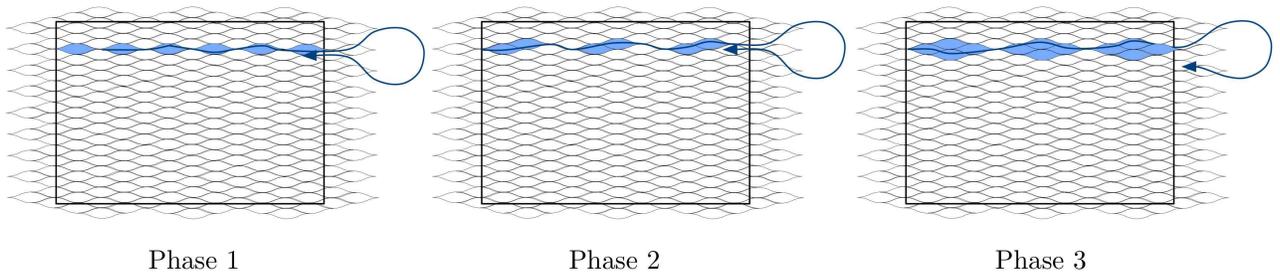

Fig. 3. Sketch of "meta-beads" at successive phases in the recursive bead tiling algorithm.

For a fixed $i \geq 1$, define a sequence of binary random variables

$$Y_t = \begin{cases} 1, & \text{if } h(t) \geq i+1 \text{ and } v_i(t-1) \leq \beta_i n; \\ 0, & \text{otherwise.} \end{cases} \quad (4)$$

In other words, $Y_t$ is equal to 1 if the $t$-th target is not visited within the first $i$ phases despite the fact that the number of beads still containing unvisited target at the inception of the $i$-th phase is less than $\beta_i n$; the values $\{\beta_i\}$ will be defined shortly.

Even though the random variable $Y_t$ depends on the targets generated before the $t$-th target, the probability that it takes the value 1 is bounded by

$$\Pr[Y_t = 1 | b(1), b(2), \ldots, b(t-1)] \leq 2^{\frac{i(i-3)}{2}} \prod_{j=1}^{i} \beta_j =: p_i,$$

regardless of the actual values of $b(1), \ldots, b(t-1)$.

It is known (e.g., see [17]) that if the random variables $Y_t$ satisfy such a condition, the sum $\sum Y_t$ is stochastically dominated by a binomially distributed random variable, namely,

$$\Pr\left[\sum_{t=1}^{n} Y_t > k\right] \leq \Pr[B(n, p_i) > k].$$

In particular,

$$\Pr\left[\sum_{t=1}^{n} Y_t > 2np_i\right] \leq \Pr[B(n, p_i) > 2np_i] < 2^{-np_i/3}, \quad (5)$$

where the last inequality is obtained using Chernoff's bound.

Let us define the sequence $\{\beta_i\}$ through the recursion

$$\beta_1 = 1,$$
$$\beta_{i+1} = 2p_i = 2^{\frac{i(i-3)}{2}+1} \prod_{j=1}^{i} \beta_j = 2^{i-2} \beta_i^2,$$

which leads to

$$\beta_i = 2^{1-i}. \quad (6)$$

With the above definition in mind, (5) can be rewritten as

$$\Pr\left[\sum_{t=1}^{n} Y_t > \beta_{i+1} n\right] \leq \Pr[B(n, p_i) > \beta_{i+1} n]$$
$$< 2^{-\beta_{i+1} n/6} = 2^{-\frac{n}{3 \cdot 2^i}}$$

which is less than $1/n^2$ for $i \leq i^*(n) := \lfloor \log_2 n - \log_2 \log_2 n - \log_2 6 \rfloor \leq \log_2 n$. Note that

$$\beta_i \leq 12 \frac{\log_2 n}{n} \quad \forall i > i^*(n). \quad (7)$$

Let $\mathcal{E}_i$ be defined as the event that $v_i(n) \leq \beta_i n$. Note that if $\mathcal{E}_i$ is true, then $v_{i+1}(n) \leq \sum_{t=1}^{n} Y_t$: the right hand side represents the number of targets that will be visited after the $i$-th phase, whereas the left hand side counts the number of beads containing such targets. We have, for all $i \leq i^*(n)$:

$$\Pr[v_{i+1} > \beta_{i+1} n | \mathcal{E}_i] \cdot \Pr[\mathcal{E}_i] \leq \Pr\left[\sum_{t=1}^{n} Y_t > \beta_{i+1} n\right] \leq \frac{1}{n^2},$$

that is,

$$\Pr[\neg \mathcal{E}_{i+1} | \mathcal{E}_i] \leq \frac{1}{n^2 \Pr[\mathcal{E}_i]},$$

and thus (recall that $\mathcal{E}_1$ is true with certainty):

$$\Pr[\neg \mathcal{E}_{i+1}] \leq \frac{1}{n^2} + \Pr[\neg \mathcal{E}_i] \leq \frac{i}{n^2}.$$

In other words, for all $i \leq i^*(n)$, $v_i(n) \leq \beta_i n$ with high probability.

Let us turn our attention to the phases such that $i > i^*(n)$. The total number of targets visited after the $i^*$-th phase is dominated by a binomial variable $B(n, 12\log_2 n/n)$; in particular,

$$\Pr[v_{i^*+1} > 24 \log_2 n | \mathcal{E}_{i^*}] \cdot \Pr[\mathcal{E}_{i^*}] \leq \Pr\left[\sum_{t=1}^{n} Y_t > 24 \log_2 n\right]$$
$$\leq \Pr[B(n, 12\log_2 n/n) > 24 \log_2 n] \leq 2^{-12 \log_2 n};$$

dealing with conditioning as before, we get

$$\Pr[v_{i^*+1} > 24 \log_2 n] \leq \frac{1}{n^{12}} + \Pr[\neg \mathcal{E}_{i^*}] \leq \frac{1}{n^{12}} + \frac{\log_2 n}{n^2}.$$

In other words, the number of targets that will be left after the $i^*$-th phase will be bounded by a logarithmic function of $n$ with high probability. ∎

### D. A bound on the length of the solution

What we know at this point is that after a sufficiently large number of phases, almost all targets will be visited, with high probability. The key point is to recognize that the length of each phase is decreasing at such a rate that the sum of the lengths of all the phases remains bounded. We first state and prove the following result which characterizes the length of Dubins' path required to execute the RECURSIVE BEAD-TILING ALGORITHM.

*Lemma 2.3:* (Length of path for the RECURSIVE BEAD-TILING ALGORITHM) Let $P \in \mathcal{P}_n$ be uniformly randomly generated in $\mathcal{Q}$. Then the length of Dubins' path required to execute $\log n$ phases of the RECURSIVE BEAD-TILING ALGORITHM over $P$ belongs to $O(n^{2/3})$.

*Proof:* Let $L_i$ denote the upper bound on the length of the path for the $i^{\text{th}}$ phase. Then one can see that

$$\sum_{i=1}^{\log_2(n)} L_i \leq 3 \sum_{j=1}^{\lceil \frac{\log_2(n)}{2} \rceil} L_{2j-1}.$$

Let us first compute the length of a pass, in either direction. The number of beads traversed will be no more than

$$\left\lceil \frac{W}{2^j l_n} \right\rceil = \left\lceil \frac{c_1}{2^j} n^{\frac{1}{3}} \right\rceil, \quad (8)$$

where $c_1 = \frac{W}{\sqrt[3]{\rho W H}}$ is a constant. The length of Dubins' path contained entirely within a meta-bead at the $(2j-1)$-th phase is less than $2^{j-1}\left(2l_n + o(l_n^2)\right)$. Hence, the total path length per pass will be bounded by:

$$\begin{aligned} L_{\text{pass},2j-1} &\leq 2^{j-1}\left(2l_n + o(l_n^2)\right)\left(\frac{c_1}{2^j}n^{\frac{1}{3}} + 1\right) \\ &= c_1 l_n n^{\frac{1}{3}} + 2^j l_n + \frac{c_1}{2}n^{\frac{1}{3}}o(l_n^2) + 2^{j-1}o(l_n^2) \end{aligned} \quad (9)$$

as $l_n \to 0^+$.

The cost of a U-turn, i.e., the length of the path needed to reverse direction and move to the next row of beads, is bounded by

$$L_{\text{u-turn},2j-1} \leq \frac{7}{3}\pi\rho + 2^{j-2}w(l_n) = \frac{7}{3}\pi\rho + 2^{j-2}\left(\frac{l_n^2}{2\rho} + o(l_n^3)\right). \quad (10)$$

The total number of passes will be at most

$$N_{\text{pass},2j-1} = \left\lceil \frac{H}{2^{j-2}w(l_n)} \right\rceil \leq \frac{\rho H}{2^{j-3}l_n^2} + 1. \quad (11)$$

The cost of closing the tour is bounded by a constant, say

$$L_{\text{closure},2j-1} \leq 4(W + H\pi\rho). \quad (12)$$

Concluding, the total path length will be bounded by

$$L_{2j-1} = N_{\text{pass},2j-1}(L_{\text{pass},2j-1} + L_{\text{u-turn},2j-1}) + L_{\text{closure},2j-1}. \quad (13)$$

Substituting eqns. (9), (10), (11) and (12) in eqn. (13), one can find constants $k_1, k_2$ and $k_3$ such that

$$\begin{aligned} L_{2j-1} &\leq k_1\left(1 + n^{\frac{2}{3}}o(n^{-\frac{2}{3}}) + n^{\frac{1}{3}}\right) + k_2 2^j\left(n^{-\frac{1}{3}} + o(n^{-\frac{2}{3}})\right) \\ &\quad + k_3 2^{-j}\left(1 + no(n^{-\frac{2}{3}}) + n^{\frac{2}{3}}\right). \end{aligned}$$

From this expression for the length of path during odd phases, one can conclude that

$$\sum_{i=1}^{\log(n)} L_i \leq 3 \sum_{j=1}^{\lceil \frac{\log(n)}{2} \rceil} L_{2j-1} \in O(n^{2/3}).$$

∎

Based on the results obtained so far, we are now ready to state an upper bound on the length of the path traveled by Dubins' vehicle to service all the $n$ targets while executing the RECURSIVE BEAD-TILING ALGORITHM followed by the ALTERNATING ALGORITHM; let $\mathrm{L}_{\text{RecBTA},\rho}(P)$ represent the corresponding quantity.

*Theorem 2.4:* (Upper bound on the length of the total path) Let $P \in \mathcal{P}_n$ be uniformly randomly generated in $\mathcal{Q}$. For all $\rho > 0$, $\mathrm{L}_{\text{RecBTA},\rho}(P) \in O(n^{2/3})$ with high probability.

*Proof:* By Lemma 2.3 the length of path to execute the RECURSIVE BEAD-TILING ALGORITHM belongs to $O(n^{2/3})$. From Theorem 2.2, the number of targets remaining at the end of the RECURSIVE BEAD-TILING ALGORITHM belongs to $O(\log n)$ with high probability. These remaining points can be serviced by any greedy policy or some heuristics (e.g., ALTERNATING ALGORITHM [9]) in $O(\log n)$ time. The statement of the theorem follows immediately. ∎

Combining results from Theorem 2.1 and Theorem 2.4, one can conclude that the RECURSIVE BEAD-TILING ALGORITHM is a constant factor approximation to the optimal DTSP with high probability.

## III. CONCLUSIONS

In this article, we have studied the TSP problem for vehicles that follow paths of bounded curvature in the plane. For the stochastic setting, we have obtained upper bounds that are within a constant factor of the lower bound established in literature [10]; the upper bounds are constructive in the sense that they are achieved by two novel algorithms. It is interesting to compare our results with the Euclidean setting (i.e., the setting in which curves do not have curvature constraints). For a given compact set and a point set $P$ of $n$ points, it is known [1], [2] that the $\mathrm{ETSP}(P)$ belongs to $\Theta(\sqrt{n})$. This is true for both stochastic and worst-case settings. In this article, we showed that, given a fixed $\rho > 0$, the *stochastic* $\mathrm{DTSP}_\rho(P)$ belongs to $\Theta(n^{2/3})$ with high probability. It is known [9] that the *worst-case* $\mathrm{DTSP}_\rho(P)$ belongs to $\Theta(n)$.

In the future, we plan to perform extensive simulations to support the results obtained in this article. Future directions of research include study of centralized and decentralized versions of the DTRP and general task assignment and surveillance problems for various non-holonomic vehicles.

## ACKNOWLEDGMENT

This material is based upon work supported in part by ONR YIP Award N00014-03-1-0512 and AFOSR MURI Award F49620-02-1-0325. The authors would like to thank John J. Enright for helpful discussions.